\documentclass[lettersize,journal]{article}

\usepackage{amsmath,amsfonts}
\usepackage{algorithmic}
\usepackage{array}
\usepackage[caption=false,font=normalsize,labelfont=sf,textfont=sf]{subfig}
\usepackage{textcomp}
\usepackage{stfloats}
\usepackage{url}
\usepackage{verbatim}
\usepackage{graphicx}
\def\BibTeX{{\rm B\kern-.05em{\sc i\kern-.025em b}\kern-.08em
    T\kern-.1667em\lower.7ex\hbox{E}\kern-.125emX}}
\usepackage{balance}
\usepackage{xcolor}
\usepackage{algorithm}
\usepackage{algorithmic}

\usepackage{pgfplots} 
\usepackage{float} 
\usepackage{caption}

\begin{document}
\title{The Unified Balance Theory of Second-Moment Exponential Scaling Optimizers in Visual Tasks}
\author{Gongyue Zhang and Honghai Liu}


\maketitle

\begin{abstract}
We have identified a potential method for unifying first-order optimizers through the use of variable Second-Moment Exponential Scaling(SMES). We begin with back propagation, addressing classic phenomena such as gradient vanishing and explosion, as well as issues related to dataset sparsity, and introduce the theory of balance in optimization. Through this theory, we suggest that SGD and adaptive optimizers can be unified under a broader inference, employing variable moving exponential scaling  to achieve a balanced approach within a generalized formula for first-order optimizers. We conducted tests on some classic datasets and networks to confirm the impact of different balance coefficients on the overall training process.

\end{abstract}


\section{Introduction}
Existing first-order optimizers mainly include two branches: classical optimizers represented by Stochastic Gradient Descent (SGD) and adaptive optimizers represented by Adam, along with their many derivatives. The debate over the merits and demerits of these two types of optimizers has persisted for a decade. In practical experience, it is generally considered that SGD is more suitable for tasks like Computer Vision(CV), while adaptive optimizers are widely used in tasks with sparse gradients, such as Large Language Models(LLM). Although adaptive optimizers always offer better convergence speeds in almost all tasks, they can lead to over-fitting in some cases, resulting in poorer generalization performance compared to SGD in certain tasks. Even in Large Language Models, Adam continues to face challenges, and its original strategy may not always have an advantage due to the introduction of improvements such as gradient clipping. With a wide variety of optimization methods available, it is essential to introduce a unified, interpretable theory. This paper will discuss under the framework of first-order optimizers and, through the intervention of the balance theory, will for the first time propose a unified strategy to integrate all first-order optimization methods.

The research motivation for this paper primarily stems from the variations in the generalization performance of optimizers due to the differences between tasks and models, which can be observed in extensive experiments. Besides the previously mentioned SGD and Adam, and their derived optimizers, which each have their specific applicability, the same optimizer can lead to different training tendencies depending on the network architecture. For example, in Wide Residual Network(WRN), when generalization performance is optimal for Cifar10 and Cifar100, there are subtle differences in the network architecture required for the same optimizer. This tendency is typically not achieved by modifying the optimizer but rather usually necessitates fine-tuning the network for different tasks. However, for models with relatively fixed network structures, or for individuals unfamiliar with the model structure, the adjustment process for different or unknown tasks can be more challenging.

This paper introduces a first-order optimizer based on the exponential scaling of second-moment, which controls the exponential coefficient to unify and derive a general formula for first-order optimizers. We also introduce balance theory to explain why this approach is effective. Finally, experiments determine the optimal balance coefficients for different groups of models and datasets. The main contributions of this paper are as follows:
\begin{itemize}
	\item{Theoretically unifies the two major optimizer branches, represented by SGD and Adam.}
	\item{Introduces a novel method for adjusting optimizers, manually intervening in training tendencies by externally estimating the degree of gradient distortion.}
	\item{Validates the sources of gradient sparsity through experiments, introduces balance theory, and enhances the interpretability of the optimization process.}
\end{itemize}

In Section 2, we introduce some foundational research relevant to our study. In Section 3, we discuss the classic phenomena of gradient vanishing and explosion to introduce our balance theory and propose a general formula for first-order optimizers based on exponential scaling of second-moment. In Section 4, we design various experiments to demonstrate the effectiveness of the balance theory.

\section{Related Work}

\subsection{Back Propagation}

Back propagation is a fundamental algorithm used for training neural networks. It operates by calculating the gradients of the loss function with respect to each weight in the network through the method of chain rule. This process starts from the output layer and proceeds backwards through the network. Each layer's output is a function of the weighted sum of the inputs it receives, which is then passed through a non-linear activation function. During back propagation, the error is calculated by comparing the predicted output of the network to the actual target values. This error is then used to calculate the gradient of the loss function for each layer, which informs how to adjust the weights to minimize the loss.

Gradient Descent is an optimization algorithm used to minimize the loss function of a model by iteratively updating the parameters, such as the weights of a neural network. The algorithm makes use of the gradients obtained from the back propagation process to adjust the parameters in the opposite direction of the gradient, aiming to reach the minimum loss. The size of the step taken in each iteration is determined by the learning rate—a crucial hyper-parameter. This process is repeated until the algorithm converges to the optimal set of parameters, ideally where the loss is minimized. Gradient descent can be performed in various forms, including batch, stochastic, and mini-batch, each differing in the amount of data used to calculate the gradient during each iteration.

\subsection{Adaptive Optimizer}

Adaptive optimizers are a class of optimization algorithms used in training machine learning models, particularly neural networks, that adjust the learning rates for each parameter individually. Unlike standard gradient descent, which uses a fixed learning rate for all parameters, adaptive optimizers modify the learning rate dynamically based on the historical gradient information of each parameter. This approach helps in addressing some of the common challenges in training, such as choosing an appropriate learning rate or dealing with the vanishing or exploding gradients.

\subsection{Gradient Clipping}

Gradient clipping is a technique used to address the problem of exploding gradients in training neural networks. Exploding gradients occur when the values of gradients become excessively large during back propagation, leading to unstable training processes where model weights receive excessively large updates and, consequently, the model fails to converge or diverges.

The core idea of gradient clipping is straightforward: it involves setting a threshold value, and if the norm (magnitude) of the gradient exceeds this threshold, the gradients are scaled down proportionally so that they stay below the threshold. This prevents any single update from being too large and helps maintain stability in the training process.

\section{Second-Moment Exponential Scaling(SMES)}

The main objective of this section is to elucidate our ideas and methods. In the first part, we discuss classic issues that have long plagued deep learning, gradient vanishing and exploding, to illustrate the sources of balancing disparities. In the second part, we point out that SGD and adaptive optimizers are essentially two special cases, and, combining balance theory, we unify the first-order optimizer approaches through a general formula involving exponential scaling of second-moment.

\subsection{Balance Theory}

In the previous section, we briefly introduced back propagation. In this section, we primarily list the classic issues encountered within back propagation. The gradient of a parameter $w$ in first-order back propagation can be expressed through the chain rule as follows:

\begin{align}
\frac{\partial L}{\partial w} = \frac{\partial L}{\partial a} \cdot \frac{\partial a}{\partial w}
\end{align}

During the back propagation process, particularly in deep neural networks, there is a loss of information in the propagation of gradients from the output layer back to the input layer. This loss of information typically manifests as either gradient vanishing or exploding, which can significantly impact the efficiency and effectiveness of network training. These issues of gradient vanishing or exploding are often related to the network's architecture and, to a lesser extent, to the complexity of the gradients in the dataset. We will explain this in two parts.

\subsubsection{Network Isomerism}

The primary sources of gradient vanishing and exploding are tied to the structure of the network. For gradient vanishing, initial discussions focused on activation functions, the depth of the network structure, and weight initialization. Among these, activation functions are one of the main factors, as the early-used sigmoid activation function has derivatives that approach zero when input values are too large or too small. Thus, similar activation functions in deep networks can lead to gradient vanishing. Additionally, in very deep networks, the cumulative product effect can also result in very small gradients. Correspondingly, gradient explosion can occur, for example, when processing long sequence data in natural language processing, where reliance on accumulations over many steps can cause substantial growth in gradients. Of course, there are now many established methods to address these issues, such as using other activation functions (such as ReLU), batch normalization, residual connections, and gradient clipping.

All methods aim to eliminate or mitigate the issues of gradient vanishing or exploding as much as possible. However, we believe that fundamentally, gradient vanishing and exploding are inherent to the process of back propagation. When different network structures are used, the degree of vanishing or exploding can vary. Here, we introduce for the first time the concept of balance, which suggests that choosing different structures within the overall network can lead to different tendencies towards balance. These tendencies accumulate during the back propagation process and consequently bias the model's overall training behavior. Selecting different network architectures will alter these tendencies, allowing the model to better adapt to the required training direction.

To further illustrate the existence of balance disparities caused by this accumulation, we can start from the most basic neural network model and consider a simple fully connected neural network. By comparing the effects of different activation functions, we can better understand how they contribute to these disparities in balance.

Consider a fully connected neural network with $L$ layers. For each layer $l$, the output $h^{l}$ is computed as:

\begin{align}
	h^{l}=\sigma(W^{l} h^{l-1}+b^{l})
\end{align}

where $W^{l}$ is the weight matrix, $b^{l}$ is the bias vector, and $\sigma$ is the activation function. The loss function is denoted as $\mathcal L$.

The gradient of the loss function with respect to the weights of the $l$-th layer $\frac{\partial\mathcal L}{\partial W^{l}}$ is computed using the chain rule:

\begin{align}
	\frac{\partial\mathcal L}{\partial W^{l}}=\frac{\partial\mathcal L}{\partial h^{L}}\prod_{k=l+1}^{L}(\frac{\partial h^{k}}{h^{k-1}})\frac{\partial h^{l}}{\partial W^{l}}
\end{align}

The term $\frac{\partial h^{k}}{h^{k-1}}$ is crucial and is affected by the activation function used:

\begin{align}
	\frac{\partial h^{k}}{h^{k-1}} = diag(\sigma'(z^{k}))W^k
\end{align}

with $z^k=W^k h^{k-1}+b^k$.

Then we can discuss impact from activation function. The Sigmoid function $\sigma(x)=\frac{1}{1+e^{-x}}$ has a derivative:

\begin{align}
	\sigma'(x)=\sigma(x)(1-\sigma(x))
\end{align}

This derivative tends to be small for values of $x$ that are not near zero (either very positive or very negative), leading to $diag(\sigma'(z^k))$ is small for large $|z^k|$. Thus, in deep networks, the product of these derivatives can result in very small gradients (gradient vanishing), especially as the number of layers increases.

The ReLU function $\sigma(x) = max(0,x)$ has a derivative:

\begin{align}
	\sigma'(x)=\begin{cases}1\;,\;if\,x\textgreater0\\0\;,\;if\,x\leq0\end{cases}
\end{align}

ReLU does not suffer from gradient vanishing in the same way as Sigmoid because the derivative is $1$ for all positive inputs. This allows gradients to pass through unaffected. However, as the depth of the network increases, ReLU may face the issue of gradient explosion when the changes in weights are significant, resulting in excessively large propagated gradients.

In summary, it is evident that different network structures result in varying tendencies in gradient propagation. Furthermore, considering that first-order optimization methods suffer from more severe information loss compared to second-order optimizers, this means that the process of back propagation is affected by an accumulation of various factors, ultimately leading to imbalances in the overall network training.

\subsubsection{Dataset Isomerism}

The composition of the dataset also impacts training. Factors such as the number of categories in the dataset, the number of samples under each category, and the characteristics of the samples themselves can influence training tendencies. For example, tasks with very simple categories but a large number of samples per category tend to be simpler for various networks. Conversely, datasets with many categories and fewer samples per category place higher demands on the classification capabilities of different networks.

From a perspective of balance, it is clear that datasets can produce different effects. The tendencies of different datasets mainly manifest as a bias towards either generalization or fitting. Datasets that typically require fitting are more prone to gradient explosion, meaning that in situations with sufficient data volume, the training process is less likely to experience gradient vanishing. On the other hand, datasets that require generalization, having fewer data points per sample, tend to lead to gradient vanishing, making it challenging to train weights near the input layer.


\subsection{Exponential Scaling}

Through the phenomena of gradient vanishing and exploding observed in back propagation, it becomes apparent that there is a balance between the tendencies towards vanishing or exploding, which primarily depends on the network architecture and other factors. In this section, we will consider the principle of balance from the perspective of optimizers, specifically how to utilize balance through fine-tuning at the optimizer level.

\subsubsection{Special Case Representation}
 
First, we consider a special case where SGD is essentially a special form of Adam. This idea is not surprising, as when the exponent of the second-moment in the denominator of an adaptive optimizer is set to zero, and without considering any other enhancements, it effectively becomes SGD itself. We can thus reformulate SGD in the form of an adaptive optimizer.

\begin{align}
	Adam\;to\;SGD:
	\theta_{t+1}=\theta_t-\frac{\eta}{(\hat{v_t})^	{0or\frac{1}{2}}+\epsilon}\cdot\hat{m_t}
\end{align}

Under this representation, SGD and adaptive optimizers can be seen as the same optimizer with separated parameters. Specifically, SGD and the default Adam correspond to exponential parameters of 0 and 0.5, respectively.

\subsubsection{Second-Moment Exponential Scaling}

Based on long-standing experience, SGD and adaptive optimizers exhibit significantly different tendencies in terms of generalization and fitting, each with its own advantages in different tasks. Adaptive optimizers generally achieve lower loss, indicating that using second-moment calculations to estimate the feature space of first-order optimizers effectively accelerates the fitting of the training set. However, they often suffer from severe over-fitting in various tasks. In contrast, in computer vision tasks, SGD tends to have a generalization advantage despite its inferior fitting capability compared to adaptive optimizers. This implies that the generalization tendency required by the task cannot be estimated solely from the training set. This means there are different optimal balance intervals for different tasks, and this information has traditionally only been obtainable by modifying the network architecture and observing the training outcomes, typically not available before training begins.

We propose a simple method that allows for balance adjustments without altering the network architecture by modifying the optimizer parameters, specifically by gradually varying the exponent to observe changes in fitting and generalization brought about by different parameters. The general formula for the scaling of the second-moment exponent(SMES) is as follows:

\begin{align}
	\theta_{t+1}=\theta_t-\frac{\eta}{(\hat{v_t})^	{\alpha}+\epsilon}\cdot\hat{m_t}
\end{align}

When the exponent hyper-parameter $\alpha$ is set to 0, it corresponds to SGD, and when it is 0.5, it corresponds to Adam. By changing the magnitude or the sign of the parameter $\alpha$, the balance of the optimizer can be altered. Using SGD as a reference, when the parameter is positive and gradually increases, the optimizer will tend to favor fitting parameters closer to the output layer. Conversely, when the parameter is negative and gradually decreases, the optimizer will tend to suppress fitting near the output layer, thereby shifting the balance towards greater generalization, or avoiding premature fitting at the output layer before adequately fitting near the input layer, which could lead to falling into saddle points.


\section{Experiment}

\subsection{Datasets \& Network}

Here, we only present the earliest experiments. We selected the Cifar-10/100 datasets and the VGG13/16/19 network datasets. The reason for choosing the VGG networks is that these networks are from early research, have simple structures, and have not been specifically adjusted for mainstream optimizers. By using these fundamental networks, we can more clearly see the impact of optimizer balance scaling coefficients on the training process.

All training sessions used a batch size of 128, an initial learning rate of 0.1, a weight decay of 5e-4, and lasted for 200 epochs. The balance coefficients were set from -0.3 to 0.1.

\subsection{Cifar-10}

We can observe that Cifar-10 is a dataset that requires significant generalization, meaning it does not require excessive fitting. When we adjust the coefficient towards the left, i.e., towards greater generalization, the generalization performance improves. This adjustment makes the network more inclined to train on the input end, allowing it to better discover the visual features of the dataset.

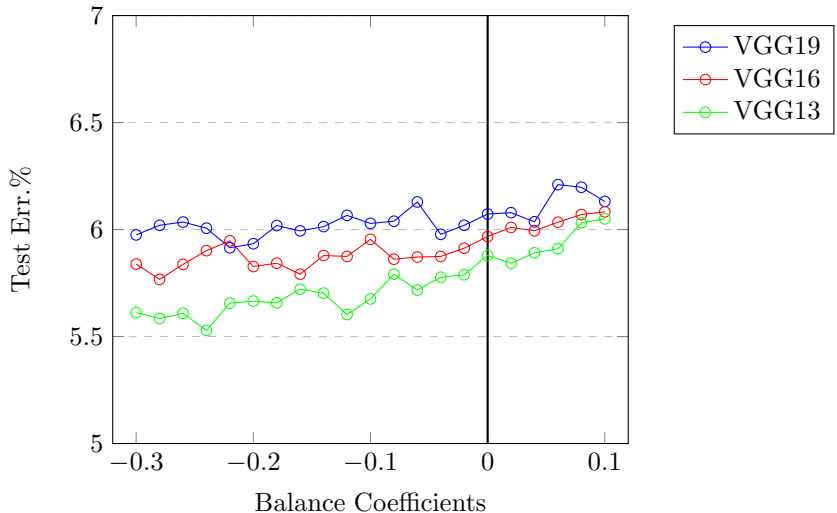
\begin{tikzpicture}
	\begin{axis}[
		xlabel={Balance Coefficients},
		ylabel={Test Err.\%},
		xmin=-0.32, xmax=0.12,
		ymin=5, ymax=7,
		ytick={5,5.5,6,6.5,7},
		legend style={at={(1.4,0.85)},anchor=east},
		ymajorgrids=true,
		grid style=dashed,
		]
		\draw [black, thick] (axis cs:0,5) -- (axis cs:0,7);
		\addplot[
		color=blue,
		mark=o,
		]
		coordinates {
(	-0.3	,	5.975	)
(	-0.28	,	6.02	)
(	-0.26	,	6.035	)
(	-0.24	,	6.00625	)
(	-0.22	,	5.915	)
(	-0.2	,	5.93375	)
(	-0.18	,	6.01875	)
(	-0.16	,	5.99375	)
(	-0.14	,	6.01375	)
(	-0.12	,	6.06625	)
(	-0.1	,	6.02875	)
(	-0.08	,	6.03875	)
(	-0.06	,	6.12875	)
(	-0.04	,	5.9775	)
(	-0.02	,	6.02	)
(	0	,	6.0725	)
(	0.02	,	6.07875	)
(	0.04	,	6.03625	)
(	0.06	,	6.21	)
(	0.08	,	6.1975	)
(	0.1	,	6.13125	)

		};
			\addlegendentry{VGG19}
		
		\addplot[
		color=red,
		mark=o,
		]
		coordinates {
(	-0.3	,	5.838571429	)
(	-0.28	,	5.767142857	)
(	-0.26	,	5.837142857	)
(	-0.24	,	5.901428571	)
(	-0.22	,	5.947142857	)
(	-0.2	,	5.827142857	)
(	-0.18	,	5.842857143	)
(	-0.16	,	5.791428571	)
(	-0.14	,	5.878571429	)
(	-0.12	,	5.874285714	)
(	-0.1	,	5.954285714	)
(	-0.08	,	5.861428571	)
(	-0.06	,	5.871428571	)
(	-0.04	,	5.874285714	)
(	-0.02	,	5.912857143	)
(	0	,	5.967142857	)
(	0.02	,	6.01	)
(	0.04	,	5.994285714	)
(	0.06	,	6.034285714	)
(	0.08	,	6.07	)
(	0.1	,	6.082857143	)

		};
			\addlegendentry{VGG16}
		
		\addplot[
		color=green,
		mark=o,
		]
		coordinates {
(	-0.3	,	5.611428571	)
(	-0.28	,	5.584285714	)
(	-0.26	,	5.608571429	)
(	-0.24	,	5.528571429	)
(	-0.22	,	5.655714286	)
(	-0.2	,	5.665714286	)
(	-0.18	,	5.657142857	)
(	-0.16	,	5.721428571	)
(	-0.14	,	5.702857143	)
(	-0.12	,	5.602857143	)
(	-0.1	,	5.675714286	)
(	-0.08	,	5.791428571	)
(	-0.06	,	5.717142857	)
(	-0.04	,	5.777142857	)
(	-0.02	,	5.788571429	)
(	0	,	5.878571429	)
(	0.02	,	5.842857143	)
(	0.04	,	5.891428571	)
(	0.06	,	5.91	)
(	0.08	,	6.031428571	)
(	0.1	,	6.05	)

		};
			\addlegendentry{VGG13}
		
	\end{axis}
\end{tikzpicture}

\subsection{Cifar-100}

We can see that for the VGG network, the Cifar-100 dataset also requires a certain degree of generalization tendency, but this tendency is not constant; there is an optimal interval. For the combination of VGG-13 and Cifar-100, the optimal point is also on the left side of SGD, with a balance coefficient of -0.08.

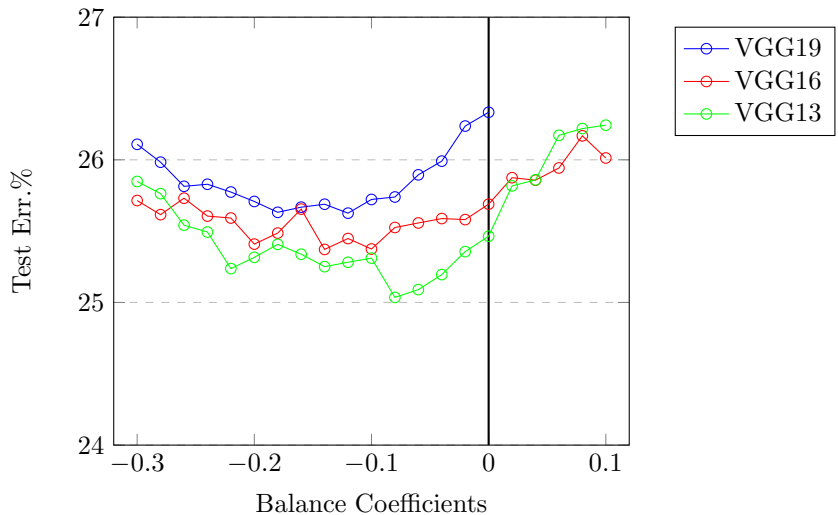
\begin{tikzpicture}
	\begin{axis}[
		xlabel={Balance Coefficients},
		ylabel={Test Err.\%},
		xmin=-0.32, xmax=0.12,
		ymin=24, ymax=27,
		ytick={24,25,26,27},
		legend style={at={(1.4,0.85)},anchor=east},
		ymajorgrids=true,
		grid style=dashed,
		]
		\draw [black, thick] (axis cs:0,24) -- (axis cs:0,27);
		\addplot[
		color=blue,
		mark=o,
		]
		coordinates {
(	-0.3	,	26.10866667	)
(	-0.28	,	25.98333333	)
(	-0.26	,	25.814	)
(	-0.24	,	25.82866667	)
(	-0.22	,	25.774	)
(	-0.2	,	25.708	)
(	-0.18	,	25.632	)
(	-0.16	,	25.668	)
(	-0.14	,	25.68866667	)
(	-0.12	,	25.626	)
(	-0.1	,	25.722	)
(	-0.08	,	25.73933333	)
(	-0.06	,	25.89466667	)
(	-0.04	,	25.99066667	)
(	-0.02	,	26.23666667	)
(	0	,	26.33333333	)
			
		};
		\addlegendentry{VGG19}
		
		\addplot[
		color=red,
		mark=o,
		]
		coordinates {
(	-0.3	,	25.71454545	)
(	-0.28	,	25.61545455	)
(	-0.26	,	25.73090909	)
(	-0.24	,	25.60545455	)
(	-0.22	,	25.59181818	)
(	-0.2	,	25.40909091	)
(	-0.18	,	25.48636364	)
(	-0.16	,	25.65454545	)
(	-0.14	,	25.37181818	)
(	-0.12	,	25.44818182	)
(	-0.1	,	25.37454545	)
(	-0.08	,	25.52545455	)
(	-0.06	,	25.55727273	)
(	-0.04	,	25.58818182	)
(	-0.02	,	25.58090909	)
(	0	,	25.68909091	)
(	0.02	,	25.87454545	)
(	0.04	,	25.85727273	)
(	0.06	,	25.94272727	)
(	0.08	,	26.16818182	)
(	0.1	,	26.01272727	)

		};
		\addlegendentry{VGG16}
		
		\addplot[
		color=green,
		mark=o,
		]
		coordinates {
(	-0.3	,	25.84818182	)
(	-0.28	,	25.76272727	)
(	-0.26	,	25.54181818	)
(	-0.24	,	25.49363636	)
(	-0.22	,	25.23727273	)
(	-0.2	,	25.31636364	)
(	-0.18	,	25.40727273	)
(	-0.16	,	25.33818182	)
(	-0.14	,	25.25090909	)
(	-0.12	,	25.28181818	)
(	-0.1	,	25.31090909	)
(	-0.08	,	25.03545455	)
(	-0.06	,	25.09	)
(	-0.04	,	25.19545455	)
(	-0.02	,	25.35636364	)
(	0	,	25.46454545	)
(	0.02	,	25.81727273	)
(	0.04	,	25.85909091	)
(	0.06	,	26.17181818	)
(	0.08	,	26.21909091	)
(	0.1	,	26.24272727	)
			
		};
		\addlegendentry{VGG13}
		
	\end{axis}
\end{tikzpicture}

\section{Conclusion}

This paper achieves the unification of first-order optimizers through the use of a variable second-moment exponent coefficient. Starting with back propagation, we introduce the balance theory of model optimization through classic phenomena such as gradient vanishing and exploding, as well as issues like dataset sparsity. Through balance theory, SGD and adaptive optimizers are unified under a broader inference, utilizing a variable scaling exponent to implement a balanced approach within the general formula for first-order optimizers. Ultimately, tests were conducted on some classic datasets and networks to confirm the impact of different balance coefficients on the overall training process.

In the future, we plan to expand more experiments and supplement theoretical details.

%

\end{document}